# Faithful Approximations of Belief Functions

**David Harmanec**
Medical Computing Laboratory
Department of Computer Science
School of Computing
National University of Singapore
Lower Kent Ridge Road
Singapore 119260

## Abstract

A conceptual foundation for approximation of belief functions is proposed and investigated. It is based on the requirements of *consistency* and *closeness*. An optimal approximation is studied. Unfortunately, the computation of the optimal approximation turns out to be intractable. Hence, various heuristic methods are proposed and experimantally evaluated both in terms of their accuracy and in terms of the speed of computation. These methods are compared to the earlier proposed approximations of belief functions.

## 1 INTRODUCTION

It is now widely accepted fact that many situations involving uncertainty cannot be very well represented within classical probability framework. Ignorance, lack of time, or a small sample size are just a few of the reasons making it hard, if not impossible, to obtain reliable point probabilities. One may then ask why the various theories of imprecise probabilities are so rarely applied to solve practical problems compared to the classical probability theory. I believe, the explanation lies in the fact that, in general, using imprecise probabilities has the computational complexity exponential in the size of the set of possible alternatives considered while using classical probabilities is only linear in the size of the universal set. Hence, the price for better accuracy or reliability seems too high. A possible solution of this dilemma is to use a subclass of imprecise probabilities that is expressive enough and at the same time inferences within this subclass are computationally tractable.

One possible candidate for such a class is the class of belief functions with the number of focal elements limited by a small constant (e.g., twice the size of the outcome space). Most algorithms involving belief functions (or their various transformations) have computational complexity polynomial in the number of focal elements. However, some algorithms may produce a belief function with a larger number of focal elements than the number of focal elements of its input belief function(s). Moreover, the original belief function may itself have more focal elements than the prescribed limit. To solve this problem we need a method that would efficiently produce ideally the best, or at least a good, approximation of a given general belief function. Addressing the development of such methods is the topic of the current paper.

This work is by no means the first attempt to develop an approximation method for belief functions. Previous work can be divided into two streams. One type of proposal was to approximate a given belief function by either a probability or possibility measure [3, 4, 17]. While it might be useful method for utilizing some of the existing techniques from probability and possibility theories on input in the form of belief function(s), it hardly qualifies as a general approximation method. The reason is that the structure of both probability and possibility measures is very simple and hence too restrictive from the point of view of belief functions. In general, it is possible to allow a more general structure of the approximating belief function while preserving the same computational complexity. Approximating belief functions by probability measures also introduces "information" that is not present in the original belief function[1]. The other direction of the research on approximation of belief functions focused on the development of a general approximation method. However, the proposals found in the literature [16, 1, 2] are not based on a well-founded approach to approximation. They are evaluated only on the basis of computer experiments using questionable measure(s) of goodness of approximation. I discuss these proposals in detail in Section 4.1. I should also mention that Monte-Carlo techniques have been applied to reasoning with belief functions [8].

---

[1] Although it is possible to preserve the amount of information carried by both the belief function and the probability measure [5].

Another approach to reduction of computational complexity found in the literature is to use a factorization of belief functions over a hypertree of sets of variables and a local computation algorithm to reason with the belief functions [7, 13, 10]. This approach is independent of using approximations and could be used in combination with it.

## 2  PRELIMINARIES

Let $X = \{x_1, x_2, \ldots, x_n\}$ denote a finite and non-empty universal set, usually referred to as a frame of discernment in the Dempster-Shafer theory [12], and let $\mathcal{P}(X)$ denote the power set of $X$.

A function $Bel : \mathcal{P}(X) \rightarrow [0,1]$, which satisfies $Bel(\emptyset) = 0$, $Bel(X) = 1$, and

$$Bel(A_1 \cup \ldots \cup A_N) \geq \sum (-1)^{|I|+1} Bel(\cap_{i \in I} A_i),$$

where the summation goes over all $\emptyset \neq I \subseteq \{1, \ldots, N\}$ for all positive integers $N$ and all $\{A_i\}_{i=1}^{N} \subseteq \mathcal{P}(X)$, is called a *belief function*.

A function $Pl : \mathcal{P}(X) \rightarrow [0,1]$ is called a *plausibility function* if it satisfies $Pl(\emptyset) = 0$, $Pl(X) = 1$, and

$$Pl(A_1 \cap \ldots \cap A_N) \leq \sum (-1)^{|I|+1} Pl(\cup_{i \in I} A_i)$$

where the summation goes over all $\emptyset \neq I \subseteq \{1, \ldots, N\}$ for all positive integers $N$ and all $\{A_i\}_{i=1}^{N} \subseteq \mathcal{P}(X)$.

A *basic probability assignment* is a function $m : \mathcal{P}(X) \rightarrow [0,1]$ such that $m(\emptyset) = 0$ and $\sum_{A \in \mathcal{P}(X)} m(A) = 1$. A subset $A$ of $X$ such that $m(A) > 0$ is called a *focal element* of $m$.

It is well known [12] that there are one-to-one correspondences between belief functions, plausibility functions and basic probability assignments. Due to these correspondences we can freely use any of them in our argumentation. Moreover, any notion associated with any of the functions $m$, $Bel$ or $Pl$ naturally translates to the remaining two by these correspondences. For example, we can talk about focal elements of a belief function meaning focal elements of its associated basic probability assignment.

In the algorithms in this paper, a basic probability assignment $m$ is represented as a set of pairs of all focal elements of $m$ and their corresponding value of basic probability assignment, i.e.,

$$m = \{(A, m(A)) \mid \emptyset \neq A \subseteq X, m(A) > 0\}.$$

If a particular set $B$ is not in any pair in $m$ then $m(B) = 0$.

## 3  OPTIMAL APPROXIMATION

In this section, I look into the problem of finding an "optimal" or "best" approximation for a given belief function. A belief function has to satisfy two basic requirements to be considered an approximation. First, it has to be a "simple" belief function.[2] In this paper, a belief function is considered simple, if it has at most $k$ focal elements, where $k$ is a small positive integer. $k$ is intentionally left unspecified as a parameter to the approximation method(s). Second, it has to be faithful. For an approximation to be faithful, the inferences made with it should be, in some sense, consistent with the inferences made with the original belief function. For an approximation to be considered best, it has to be the approximation that is the closest to the original. The next two subsections make these notions more precise.

### 3.1  CONSISTENCY OF BELIEF FUNCTIONS

The faithfulness of the approximation is made precise by the notion of consistency.

**Definition 1** *Let $Bel$ and $Bel'$ denote two belief functions on $X$. We say that $Bel'$ is (weakly) consistent with $Bel$ if and only if $Bel(A) \geq Bel'(A)$ for all $A \subseteq X$.*

The definition states that $Bel'$ is consistent with $Bel$ if it does not ascribe a larger belief than $Bel$ to any subset of $X$, i.e., $Bel$ is more precise — the interval $[Bel'(A), Pl'(A)]$ contains the interval $[Bel(A), Pl(A)]$ for each $A \subseteq X$. It appears very natural to ask that an approximation is consistent with the original belief function, no matter what interpretation of belief functions is used. Some authors (e.g.,[4]) use the name (weak) inclusion instead of consistence for the same property. I prefer the name consistence as it is not tied to the random set interpretation of belief functions.

Although the notion of consistence is very natural, it does not offer a guidance how to actually compute an approximation of a given belief function. The notion of strong consistency, though admittedly less intuitive, offers such a guidance.

---

[2] I could consider "approximations" by arbitrarily complex belief functions, but as I am interested in reducing computational complexity, I explicitly exclude complex belief functions from the class of approximations.

**Definition 2** *Let Bel, Bel' denote two belief function on $X$ and let $m$, $m'$ denote their corresponding basic probability assignments. Moreover, let $A_1, A_2, \ldots, A_p$ ($p \geq 1$) denote all focal elements of $m$ and let $B_1, B_2, \ldots, B_q$ ($q \geq 1$) denote all focal elements of $m'$. We say that Bel' is strongly consistent with Bel if and only if there is a collection of numbers $w_{ij}$, $i = 1, 2, \ldots, p$, $j = 1, 2, \ldots, q$, such that*

$$\sum_{i=1}^{p} w_{ij} = m'(B_j), \text{ for all } j,$$

$$\sum_{j=1}^{q} w_{ij} = m(A_i), \text{ for all } i,$$

*and $B_j \not\subseteq A_i \implies w_{ij} = 0$.*

A belief function $Bel'$ is strongly consistent with another belief function $Bel$ if the basic probability assignment value of every focal element of $Bel'$ is a sum of fractions of basic probability assignment values of some focal elements of $Bel$ that are its subsets. In other words, (some fractions of) basic probability values of focal elements of the "original" belief function are moved to a superset in the "approximation". As expected, the strong consistence implies consistence; the inverse does not hold in general [3].

Now we can define precisely what is meant by "approximation" in this paper.

**Definition 3** *A belief function $Bel'$ is called (strong) $k$-approximation of a given belief function $Bel$ if $Bel'$ is (strongly) consistent with $Bel$ and $Bel'$ has at most $k$ focal elements.*

### 3.2  MEASURE OF CLOSENESS

To measure the "closeness" of an approximation, I use the function $DF_{Bel}$ defined by

$$DF_{Bel}(Bel') = \sum_{A \in \mathcal{P}(X)} (Bel(A) - Bel'(A))$$

for any belief function $Bel'$ on $X$ consistent with $Bel$. To obtain the best approximation we need to minimize this function over all (strong) $k$-approximations of $Bel$. That is, we try to minimize the sum of differences of a given belief function and the "approximating" belief function.

### 3.3  AN ALGORITHM FOR COMPUTING AN OPTIMAL APPROXIMATION

In the rest of the paper, I investigate only strong $k$-approximations of a given belief function. However, if the following conjecture holds, finding optimal $k$-approximation is the same thing as finding optimal strong $k$-approximation.

**Conjecture 1** *Let $Bel$ denote a belief function on $X$. Any belief function $Bel'$ that is consistent with $Bel$ and minimizes $DF_{Bel}$ among all the belief functions consistent with $Bel$ is also strongly consistent with $Bel$.*

Our goal is to find optimal strong $k$-approximation of a given belief function $Bel$. In general, there may be many optimal strong $k$-approximations. We want to find at least one. The theorem below suggests where to start looking.

**Theorem 1** *Let $Bel$ denote a belief function on $X$ and let $Bel'$ denote a strong $k$-approximation of $Bel$. Then there exists a strong $k$-approximation $Bel''$ of $Bel$ such that its focal elements are also focal elements of $Bel'$, (under the notation of Definition 2) $w_{ij}^{Bel''} \in \{0, m(A_i)\}$ and $DF_{Bel}(Bel') \geq DF_{Bel}(Bel'')$.*

As a consequence of the above theorem, we only need to explore the partitions of the set of focal elements of the original belief function into $k$ parts to find the optimal strong $k$-approximation. The following algorithm is a formalization of this idea.

**Algorithm 1 (Optimal approximation)**
*Input:* a basic probability assignment
   $M = \{\langle B_i, m(B_i)\rangle \mid i = 1, 2, \ldots, s\}$, number of
   focal elements for the approximation $k$
*Output:* optimal approximation of $M$
**if** $s \leq k$ **then**
   return $M$
**else**
   Find the first partition $\mathcal{K} = \{P_1, P_2, \ldots, P_k\}$
   of $\{1, 2, \ldots, s\}$
   $OPT = \{\langle \bigcup_{j \in P_l} B_j, \sum_{j \in P_l} m(B_j)\rangle \mid P_l \in \mathcal{K}, l = 1, 2, \ldots, k\}$
   **while** there is an unexamined partition
   of $\{1, 2, \ldots, s\}$ **do**
     Find the next partition $\mathcal{K} = \{P_1, P_2, \ldots, P_k\}$
     of $\{1, 2, \ldots, s\}$
     $WM = \{\langle \bigcup_{j \in P_l} B_j, \sum_{j \in P_l} m(B_j)\rangle \mid P_l \in \mathcal{K}, l = 1, 2, \ldots, k\}$
     **if** $DF(WM) < DF(OPT)$ **then**
       $OPT = WM$
     **end if**
   **end while**
   return $OPT$
**end if**

Unfortunately, going through all possible partitions is not computationally tractable.

## 4   HEURISTIC APPROXIMATIONS

As seen in the previous section, it is not computationally feasible to find an optimal approximation of a given belief function. As usual in such a situation, one needs to turn to a heuristic method that does not guarantee optimal results but empirically provides good approximations. In this section, I propose several such heuristic methods. I also overview and discuss proposals from the literature for heuristic approximation methods. The experimental evaluation of these methods is the topic of the next section.

### 4.1   PREVIOUS WORK

This subsection is an overview of the previous proposals for approximations of belief functions. As all the proposals are heuristic methods, I discuss them within a section on heuristic algorithms.

Voorbraak [17] proposed *Bayesian approximation* of belief functions that approximates given belief function by a probability measure. It is well known [12] that a probability measure is a special type of belief function with only singleton focal elements. For a given basic probability assignment $m$ the basic probability assignment $m'$ of the Bayesian approximation of $m$ is given by

$$m'(A) = \begin{cases} \frac{\sum_{B | A \subseteq B} m(B)}{\sum_{C | C \subseteq X} m(C) \cdot |C|}, & |A| = 1 \\ 0, & \text{otherwise.} \end{cases}$$

Obviously the Bayesian approximation has at most $|X|$ focal elements. Unfortunately, the Bayesian approximation is generally not consistent with the original belief function, which is one of our fundamental requirements.[3] Bayesian approximation commutes with the Dempster rule of combination, which is its distinguishing characteristic.

In my opinion, the most well founded approximation method found in the literature is the *consonant approximation* of a belief function proposed by Dubois and Prade [4]. Their goal is to find a maximal consonant strongly consistent belief function that maximizes imprecision (i.e., the expression $\sum m(A) \cdot |A|$). This is an NP-hard problem so they proposed a heuristic algorithm as a computationally tractable approximation. Like the Bayesian approximation, the consonant approximation has at most $|X|$ focal elements. As mentioned above, the consonant approximation is a faithful approximation. Its main limitation is its restriction to consonant belief functions. See Section 5

---

[3]This is true for any "approximation" by a probability measure, which also excludes using Smet's *pignistic probability* as an approximation [14].

for quantitative comparison with other faithful heuristic methods.

Lowrance et al [9] appear to be the first to propose a general method, called *summarization*, for approximating belief functions. Let $m$ denote a given basic probability assignment with $s$ focal elements $A_1, A_2, \ldots, A_s$ ordered in such a way that $m(A_i) \geq m(A_{i+1})$ for all $i \in \{1, 2, \ldots, s-1\}$. Then the summarization approximation $m'$ is given by

$$m'(A) = \begin{cases} m(A), & A = A_i, i \in \{1, \ldots, k-1\} \\ \sum_{j=k}^{s} m(A_j) & A = \bigcup_{j=k}^{s} A_j \\ 0, & \text{otherwise} \end{cases} \quad (1)$$

This approximation is also strongly consistent with the original belief function.

The first systematic study of approximations of belief functions was done by Tessem [16]. He also proposed a new method called $k$-$l$-$x$ approximation. The $k$-$l$-$x$ approximation works by preserving the original focal elements with the highest basic probability assignment values and renormalizing the values. Unfortunately, the resulting belief function is not, in general, consistent with the original belief function. Tessem uses the maximum difference over all subsets of $X$ between the pignistic probability [14] corresponding to the original and approximating belief function as the measure of closeness (or error measure in his terminology). The use of this measure is questionable. There are other possible 'representations' of a belief function by a probability measure (e.g., maximum entropy probability) and one can also use the belief function directly in decision making [6, 11, 15, 18].

Bauer [1, 2] did a second, and, to my knowledge, last systematic study of approximations of belief functions. He also proposed a new approximation method, called $D1$. The $D1$ approximation works by keeping $k-1$ focal elements from the original basic probability assignment and distributing the mass from the rest of the focal elements to these or to $X$. The distribution is done either to the minimal supersets if any, or to minimal non-disjoint sets with larger cardinality if any (according to the proportion of their intersection with the distributed focal element), or to $X$ as the last resort. Again, the $D1$ approximation method does not always produce a belief function that is consistent with the original belief function. Bauer uses the main Tessem's error measure as well as two new ones. These measures suffer from the same drawbacks as the main Tessem's error measure as they are also based on the pignistic probability. Moreover, the motivation of these measures is dubious. The author seems to suggest that the user would make the decision on the basis of the highest pignistic probability. How-ever, this is not the standard decision making set up, when the user has a choice of actions and makes a decision based on utility expectations with respect to probabilities (belief functions) conditional on taking a particular action.

## 4.2  PAIR APPROXIMATION

The first heuristic approximation method is a variation of the standard one-step-look-ahead (or greedy) heuristic. It reduces the number of focal elements of the current basic probability assignment by one at each step (starting from the original) until the desired number $k$ of focal elements is reached. The reduction is done by merging the two focal elements, merging of which results in the smallest $DF_{Bel}$, where $Bel$ is the belief function corresponding to the current basic probability assignment. The following result is used in the selection.

**Proposition 1** *Let $m$ denote a basic probability assignment on $X$ and $A$, $B$ denote two different focal elements of $m$. Let $m'$ denote a basic probability assignment obtained from $m$ by merging $A$ and $B$, i.e.,*

$$m'(C) = \begin{cases} 0, & C \in \{A,B\} \\ m(A) + m(B) & C = A \cup B \\ m(C) & \text{otherwise} \end{cases}$$

*Then*

$$DF_{Bel}(Bel') = m(A) \cdot 2^{|X-A|} + m(B) \cdot 2^{|X-B|} \quad (2)$$
$$- (m(A) + m(B)) \cdot 2^{|X-A\cup B|},$$

*where $Bel$ and $Bel'$ denote, respectively, the belief function corresponding to $m$ and $m'$.*

Let $CP(A,B)$ denote the expression on the left hand side of (2). The algorithm is presented below.

**Algorithm 2 (Pair approximation)**
**Input:**  a basic probability assignment
  $M = \{\langle B_i, m(B_i) \rangle \mid i = 1,2,\ldots,s\}$, number of focal elements for the approximation $k$
**Output:**  Pair approximation of $M$
  **if** $s \leq k$ **then**
    **return** $M$
  **else**
    $WM = M$
    **for** $\langle A, m(A) \rangle \in WM$ **do**
      $C_A^{WM} =$
        $\min_{B|\langle B,m(B)\rangle \in \{WM - \{\langle A,m(A)\rangle\}\}} CP(A,B)$
    **end for**
    **while** $|WM| > k$ **do**
      $B = \arg\min_{A|\langle A,m(A)\rangle \in WM} C_A^{WM}$
      $C =$
        $\arg\min_{A|\langle A,m(A)\rangle \in \{WM-\{\langle B,m(B)\rangle\}\}} CP(A,B)$
      $WM = (WM -$
        $\{\langle B,m(B)\rangle, \langle C,m(C)\rangle, \langle B\cup C, m(B\cup C)\rangle\}) \cup$
        $\{\langle B\cup C, m(B) + m(C) + m(B\cup C)\rangle\}$
      recompute $C_A^{WM}$
    **end while**
    **return** $WM$
  **end if**

## 4.3  SINGLE APPROXIMATION

The Pair approximation is still quite computationally complex. It has the worst case computational complexity $O(s^3)$. This is due to the fact that the cost of merging is associated with pairs of focal elements. To reduce the complexity one could try to associate a "cost" with the individual focal elements instead of their pairs, and, at each step, merge the two focal elements with the lowest "cost". This is what the Single approximation is doing. Here, the amount of increase of $DF$ by merging a focal element with $X$ is taken as the "cost". From (2) we have

$$CS(A) = m(A) \cdot 2^{|X-A|} + m(X) \cdot 2^{|X-X|}$$
$$- (m(A) + m(X)) \cdot 2^{|X-A\cup X|}$$
$$= m(A) \cdot \left(2^{|X-A|} - 1\right).$$

The algorithm is presented below.

**Algorithm 3 (Single approximation)**
**Input:**  a basic probability assignment
  $M = \{\langle B_i, m(B_i) \rangle \mid i = 1,2,\ldots,s\}$, number of focal elements for the approximation $k$
**Output:**  Single approximation of $M$
  **if** $s \leq k$ **then**
    **return** $M$
  **else**
    $WM = M$
    **while** $|WM| > k$ **do**
      $B = \arg\min_{A|\langle A,m(A)\rangle \in WM} CS(A)$
      $C =$
        $\arg\min_{A|\langle A,m(A)\rangle \in \{WM-\{\langle B,m(B)\rangle\}\}} CS(A)$
      $WM = (WM -$
        $\{\langle B,m(B)\rangle, \langle C,m(C)\rangle, \langle B\cup C, m(B\cup C)\rangle\}) \cup$
        $\{\langle B\cup C, m(B) + m(C) + m(B\cup C)\rangle\}$
    **end while**
    **return** $WM$
  **end if**

## 4.4  RATIO APPROXIMATION

The choice of $X$ as the foci of the merging in the Single approximation is somewhat arbitrary. The Ratio approximation goes to the other extreme and computes the "cost" based on the assumption of merging with a superset containing just one extra focal element.

Again, from (2) we have

$$CR'(A) = m(A) \cdot 2^{|X-A|} + m(A \cup \{\alpha\}) \cdot 2^{|X-A\cup\{\alpha\}|}$$
$$- (m(A) + m(X)) \cdot 2^{|X-A\cup A\cup\{\alpha\}|}$$
$$= m(A) \cdot \frac{2^{|X|-1}}{2^{|A|}}.$$

As we are interested only in the comparison of the values of $CR'$ and $CR'(A) < CR'(B)$ if and only if $\frac{m(A)}{2^{|A|}} < \frac{m(B)}{2^{|B|}}$, it is simpler to use

$$CR(A) = \frac{m(A)}{2^{|A|}}$$

instead of $CR'$. The algorithm of the Ratio approximation is the same as the algorithm of the Single approximation with $CR$ replacing $CS$.

Obviously, one could interpolate between the two extremes presented by the Single and Ratio approximations and select a superset with any number of focal elements between $|A|+1$ and $|X|$ as the basis of the heuristic "cost". One could even try to select it on the basis of the average size of focal elements in the original. Or, alternatively, one could try to drop using supersets at all. However, I have not pursued these ideas further (yet).

### 4.5 LUMP APPROXIMATION

If even the speed of the Single or Ratio approximations is not acceptable, one can use similar idea to that of the summarization method given by (1) and merge all $|X|+1-k$ focal elements with the smallest "cost" in one step instead of doing it iteratively. I call the resulting method the Lump approximation. Let $m$ denote a given basic probability assignment with $s$ focal elements $A_1, A_2, \ldots, A_s$ ordered in such a way that $CS(A_i) \geq CS(A_{i+1})$ (or $CR(A_i) \geq CR(A_{i+1})$ or any other "cost")[4] for all $i \in \{1, 2, \ldots, s-1\}$. Then the Lump approximation $m'$ is given by

$$m'(A) = \begin{cases} m(A), & A = A_i, i \in \{1, \ldots, k-1\} \\ \sum_{j=k}^{s} m(A_j) & A = \bigcup_{j=k}^{s} A_j \\ 0, & \text{otherwise} \end{cases}$$

### 4.6 ITERATIVE APPROXIMATION

The last method I considered is a variation on the "greatest descent" theme. The method starts from a random partition of $s$ focal elements into $k$ parts. The partition corresponds to a belief function consistent with the original belief function the same way as in the case of the optimal approximation. It then iteratively improves (hence the name Iterative approximation) its current approximation by looking at the neighboring partition until no improvement is possible or the specified time or number of iterations is exceeded.[5] A partition is neighboring if it can be obtained from the given partition by moving one element from one part to a different part. The exact algorithm follows.

**Algorithm 4 (Iterative approximation)**
*Input:* a basic probability assignment
  $M = \{\langle B_i, m(B_i)\rangle \mid i = 1, 2, \ldots, s\}$, number of focal elements for the approximation $k$,
  max.running time $Time$, max. number of iterations $NumIt$
*Output:* Pair approximation of $M$
  **if** $s \leq k$ **then**
    **return** $M$
  **else**
    Find a random partition $\mathcal{BK} = \{P_1, P_2, \ldots, P_k\}$ of $\{1, 2, \ldots, s\}$
    $\mathcal{BM} = \{\langle \bigcup_{j \in P_l} B_j, \sum_{j \in P_l} m(B_j)\rangle \mid P_l \in \mathcal{BK}, \, l = 1, 2, \ldots, k\}$
    $NumIt = 1$
    **while** $Time$ and $NumIt$ is not exceeded **do**
      $\mathcal{KM} = \mathcal{BM}$
      $\mathcal{KK} = \mathcal{BK}$
      **for** $p \in \{1, 2, \ldots, k\}$, $q \in \{1, 2, \ldots, k\}$, $p \neq q$, $|P_p| > 1$ **do**
        **for** $A \in P_p$ **do**
          $\mathcal{K} = (\{P_1, P_2, \ldots, P_k\} - \{P_p, P_q\}) \cup \{P_p - \{A\}, P_q \cup \{A\}\}$
          $\mathcal{WM} = \{\langle \bigcup_{j \in P_l} B_j, \sum_{j \in P_l} m(B_j)\rangle \mid P_l \in \mathcal{K}, \, l = 1, 2, \ldots, k\}$
          **if** $DF(\mathcal{WM}) < DF(\mathcal{BM})$ **then**
            $\mathcal{BM} = \mathcal{WM}$
            $\mathcal{BK} = \mathcal{K}$
          **end if**
        **end for**
      **end for**
      **if** $DF(\mathcal{KM}) = DF(\mathcal{BM})$ **then**
        **return** $\mathcal{BM}$
      **end if**
      $NumIt = NumIt + 1$
    **end while**
    **return** $\mathcal{BM}$
  **end if**

## 5 EXPERIMENTS

In this section I report on the experiments I conducted to evaluate the accuracy and speed of the computation of the heuristic approximation methods I am proposing in this paper as well as the Consonant and Summarization approximations (the only earlier proposals that are faithful). The experimental framework was implemented in C++. Most of the experiments were conducted on a Sun Ultra 1 workstation running Solaris 2.5, but some were also run on a 166Mhz Pentium MMX PC running Windows 95 to eliminate the possibility of dependency of the results on the computing platform.

---

[4]In my experiments I used $CS$ as the "cost" in the Lump approximation.

[5]I did not limit the time or the number of iterations in the experiments.

In each run of an experiment, I have randomly generated 1000 basic probability assignments of a given specification and computed the various heuristic approximations recording their average scaled $DF$ from the original as well as the average actual time it took to compute them. The minimum and maximum $DF$ of a consistent belief function depends on the original belief function. To get comparable results, I computed the optimal approximation and scaled the $DF$ values of the individual heuristic approximation methods between the $DF$ of the optimal approximation and the $DF$ of the vacuous belief function (which is trivially consistent) for each original belief function. A huge limitation of this approach is the need to compute the optimal approximation. This can be done only for belief functions with very small number of focal elements. I was able to do it for belief functions with up to 15 focal elements. In another set of experiments I scaled the $DF$ values only by the $DF$ of the vacuous belief function. Unlike Tessem [16] and Bauer [1, 2], I generated the basic probability assignments uniformly random as I see no compelling reason to do otherwise. To be able to compare the other approximation methods with the Consonant approximation, I used $k = |X|$ in all the experiments, as this requirement is hard-wired into the Consonant approximation. I did the experiments for $|X| \in \{3, 4, \ldots, 8\}$ and for a particular fixed $|X|$ I generated originals with $|X|+1, |X|+2, \ldots, 2^{|X|} - 1$ focal elements. The experiments seem to suggest that the Summarization approximation method is the fastest one to compute but the least (or second least) accurate. The Pair method is the most accurate but the second slowest after the Iterative method. The Iterative method is the only clearly dominated method as for larger number of focal elements it both takes the longest time to compute[6] and still is the least (or second least) accurate. A sample of typical results of the experiments can be found at Figure 1 and Table 1.

---
[6]Of course, this could be improved by restricting the number of iterations or allocated time. But that would mean further reducing the accuracy of the method.

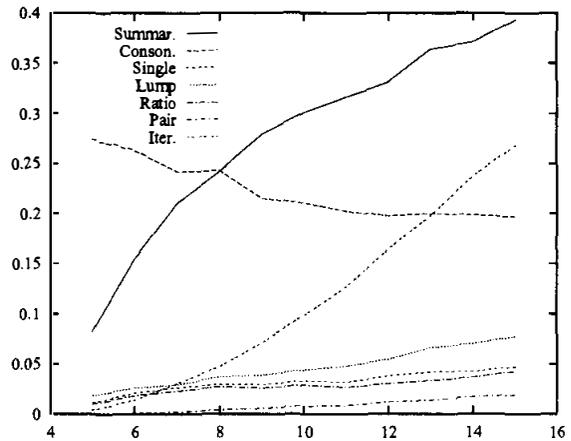

Figure 1: Sample Average Scaled $DF$ of Approximations ($|X| = 4$)

## 6   CONCLUSIONS

In this paper, I propose a foundation for approximations of belief functions. Besides simplicity, consistency with the original and closeness to the original are viewed as basic requirements of an approximation. An algorithm for computing optimal approximation is developed. As the algorithm is not computationally feasible, I suggest several heuristic methods that can be used in practical situations. I also report the results of an experimental evaluation of several heuristic approximation methods both suggested in this paper and found in the literature. The results show that the Pair approximation is on average the closest to the original from the methods considered and the Summarization approximation of Lowrance et al [9] takes the shortest time to compute. Somewhat surprisingly and contrary to the previous studies [16, 1, 2] the Consonant approximation of Dubois and Prade [4] does not perform as badly as one might expect. It is my hope that the results presented here will help to make belief functions a *practical* alternative to precise probability as a tool for dealing with uncertainty.

### Acknowledgments

The work on this paper has been supported by a Strategic Research Grant No. RP960351 from the National Science and Technology Board and the Ministry of Education, Singapore.

Table 1: Sample Average Running Time of Approximations in Seconds ($|X| = 4$)

| s  | Summar. | Consonant | Iterative | Lump    | Pair    | Ratio   | Single  |
|----|---------|-----------|-----------|---------|---------|---------|---------|
| 5  | 0.00011 | 0.00035   | 0.01076   | 0.00033 | 0.00104 | 0.00023 | 0.0001  |
| 6  | 0.00012 | 0.00032   | 0.0276    | 0.00039 | 0.00196 | 0.00041 | 0.00044 |
| 7  | 0.00008 | 0.00041   | 0.04837   | 0.00056 | 0.00355 | 0.00072 | 0.00054 |
| 8  | 0.00016 | 0.00039   | 0.07394   | 0.00078 | 0.00511 | 0.00078 | 0.00096 |
| 9  | 0.00021 | 0.00054   | 0.10779   | 0.00097 | 0.00723 | 0.00115 | 0.00116 |
| 10 | 0.00011 | 0.00069   | 0.14258   | 0.00104 | 0.00943 | 0.0016  | 0.00138 |
| 11 | 0.00009 | 0.00062   | 0.18435   | 0.00126 | 0.01207 | 0.00189 | 0.00182 |
| 12 | 0.00018 | 0.00075   | 0.22368   | 0.00162 | 0.01509 | 0.00194 | 0.00208 |
| 13 | 0.00019 | 0.00068   | 0.27124   | 0.00184 | 0.01861 | 0.00228 | 0.00215 |
| 14 | 0.00024 | 0.0008    | 0.31646   | 0.00192 | 0.02239 | 0.00266 | 0.00251 |
| 15 | 0.0002  | 0.00067   | 0.38495   | 0.00216 | 0.0275  | 0.00289 | 0.00297 |